%% file: root.tex
\documentclass[letterpaper, 10 pt, conference]{ieeeconf}  

\IEEEoverridecommandlockouts                              

\usepackage{graphics} 
\usepackage{epsfig} 
\usepackage{times} 
\usepackage{amsmath} 
\usepackage{amssymb}  
\usepackage{makecell}
\usepackage{multirow}
\usepackage{hyperref}

\usepackage{mathtools}
\usepackage{mathrsfs}
\usepackage{hyperref}
\usepackage{float}
\usepackage{pifont}
\newcommand{\cmark}{\ding{51}}%
\newcommand{\xmark}{\ding{55}}%
\usepackage{xcolor}
\usepackage{booktabs} 
\usepackage{graphicx}
\usepackage{threeparttable} 
\usepackage{tabularx}
\usepackage{gensymb}
\usepackage{afterpage}
\usepackage{caption}
\newcommand\blfootnote[1]{%
  \begingroup
  \renewcommand\thefootnote{}%
  \footnote{#1}%
  \addtocounter{footnote}{-1}%
  \endgroup
}
\usepackage[abs]{overpic}  

\usepackage{fancyhdr}
\fancypagestyle{firstpage}{ 
    \fancyhf{} 
    \fancyhead[L]{Accepted to IEEE International Conference on Robotics and Automation (ICRA) 2026} 
}

\title{\LARGE \bf
 Overcoming Imperfect Kinematics in Surgical Robotics Through Sim-to-Real Visuomotor Learning
}

\author{
\authorblockN{
        Zhaoxuan Yan$^{1,2}$,
        Kaizhong Deng$^{1,2}$,
        Zhaoyang Jacopo Hu$^{1,3}$,\\
        George P. Mylonas$^{1,2}$\IEEEmembership{Member, IEEE},
        Daniel S. Elson$^{*1,2}$
        }
}

\begin{document}
\bstctlcite{IEEEexample:BSTcontrol}

\input{figures/graphical_abstract/graphical_abstract}

\thispagestyle{firstpage}
\pagestyle{empty}


\begin{abstract}

Robot-Assisted Surgery is integral to modern minimally invasive procedures, with automation emerging as the next frontier to enhance precision and reduce surgeon fatigue. This evolution is largely impeded by the inherent kinematic inaccuracies of surgical robots, where unreliable internal sensors lead to significant control errors. While previous methods attempted to mitigate these issues through complex model-based calibration, they often suffer from high cost and limited effectiveness. This work utilises a learning-policy to actively compensate for hardware inaccuracies using closed-loop visual feedback that was trained from a teacher-student learning framework. The policy can fuse unreliable internal readings with precise external visual data, allowing it to correct for kinematic errors in real time without needing a perfect physical model. The learned policy was successfully deployed on the da Vinci Research Kit, where experiments validated the fundamental feasibility of using external vision to overcome internal sensor deficits. This research provides a foundational and reliable control methodology, paving the way for more advanced and robust surgical automation.

\end{abstract}

\blfootnote{
$^{1}$Hamlyn Centre for Robotic Surgery, Institute of Global Health Innovation, Imperial College London, Exhibition Road, London, SW7 2AZ, UK.

$^{2}$Department of Surgery and Cancer, Imperial College London.

$^{3}$Department of Mechanical Engineering, Imperial College London.

$^{*}$Corresponding author. 

Corresponding email: {\tt\small daniel.elson@imperial.ac.uk}. 

More information can be found on the \href{https://kdeng98.github.io/about_me/projects/icra26-dvrk/index.html}{project page}
}

\section{INTRODUCTION}

Robot-Assisted Surgery has gained widespread application in the field of minimally invasive surgery, with systems like the da Vinci surgical system becoming a benchmark of commercial and clinical success~\cite{surgical_review}. 
Building on this foundation, the field is advancing from teleoperation towards task automation, aiming to develop intelligent systems capable of autonomously executing surgical procedures that transforms the robot from a passive tool into an active, intelligent agent~\cite{review_auto_surgery}.

Previous studies with a focus of surgical automation have attempted peg transfer~\cite{Auto_peg_transfer1, Auto_peg_transfer2, Auto_peg_transfer_3}, shunt insertion \cite{stunt1, stunt2}, endoscopic control~\cite{Auto_endoscope}, instrument manipulation~\cite{Auto_instrument_mani}, tissue manipulation~\cite{Auto_tissue_mani}, tissue retraction~\cite{Auto_tissue_retraction}, and tissue resection~\cite{Auto_retraction_1}. It is worth noting that some of the methodology designs are shifting from task-specific solutions towards being general-purpose task-agnostic agents~\cite{SRT-H}.

The foundation of such a task-agnostic approach is the ability to reliably execute any given trajectory~\cite{VPPV}. However, it is fundamentally challenged by the surgical system's inherent kinematic inaccuracy~\cite{kim2024srt}. This inaccuracy — arising from hardware factors such as hysteresis, mechanical flexibility, and joint clearance — yields unreliable joint angle measurements~\cite{kinematic_inaccuracy}. Consequently, a discrepancy emerges between the robot's commanded trajectory and its actual executed motion, affecting the robot's ability to perform delicate surgical tasks. Addressing this challenge remains essential for reliable surgical automation.

Classical approaches typically rely on model-based calibration to improve kinematic accuracy. Geometric calibration adjusts Denavit-Hartenberg (DH) parameters by aligning measured and modelled end-effector poses~\cite{DH}; gravity compensation identifies dynamic parameters to offset gravitational and cable-induced disturbances~\cite{gravity_compensation}; and vision-based calibration employs RGB-D sensing to correct cable-driven inaccuracies~\cite{vision_compensate}. However, such methods generalise poorly across surgical environments and offer limited overall effectiveness~\cite{model_limitations}. More recently, learning-based approaches have shown promise in estimating and compensating for kinematic errors~\cite{Kinematic_error_learning_based_estimator, Kinematic_error_learning_based_compensator}, yet typically require precise end-effector pose measurements, which are difficult to obtain in practice.

Recent work~\cite{ha2024umilegs, wu2024helpful} decouples the policy architecture into a high-level planner and a low-level controller. This separation enables task-level reasoning independent of complex robot dynamics~\cite{fu2024humanplus}. Therefore, the low-level controller can focus on a task-agnostic trajectory tracking~\cite{fu2022deep}. Our work targets this low-level controller, as reliable trajectory execution is a prerequisite for the success of any high-level planner.

While the hierarchical structure simplifies the controller's objective, training a low-level policy is challenging due to the cable-drive surgical robot's unreliable proprioceptive data. The teacher-student paradigm is an effective methodology across various complex robots, including humanoids~\cite{shao2025langwbclanguagedirectedhumanoidwholebody}, quadrupeds~\cite{mousa2025tarteacheralignedrepresentationscontrastive}, and manipulators~\cite{ankile2024imitationrefinementresidual}.
In this framework, a teacher policy is first trained in simulation with access to privileged information, such as accurate physical state. It is then distilled into a student policy that relies solely on the imperfect sensory data available on the physical system. This naturally frames kinematic inaccuracy as a discrepancy between an ideal teacher with a perfect kinematic model and a realistic student that must operate with the faulty kinematics

This paper proposes a learning-based teacher-student framework that leverages closed-loop visual feedback to actively compensate for the system's inherent kinematic errors. The approach has been validated in high-fidelity simulation~\cite{yu2024orbit} and subsequently deployed onto a physical robotic platform, da Vinci Research Kit (dVRK)~\cite{dVRK}, via sim-to-real transfer.
The main contributions of this work are:
\begin{itemize}
    \item A low-level visual–motor controller, trained in a teacher–student framework, that learns to compensate for internal kinematic errors using external visual feedback.
    \item A sim-to-real transfer evaluation on a physical dVRK platform in a serial setup, demonstrating competitive performance compared to both classical control methods and learning-based policies.
    \item Real-world experiments that highlight the robustness of the policy to the dVRK’s kinematic inaccuracies, establishing a foundation for future research in surgical automation.
\end{itemize}

\section{METHOD}

\subsection{Overview of Teacher Student Framework}

As illustrated in \autoref{fig:pipeline}, our methodology is centered on a teacher-student learning framework designed for sim-to-real transfer. The teacher-student framework comprises two distinct training phases designed to address the challenges of precise trajectory tracking in robotic surgery. Initially, the teacher policy is trained using Reinforcement Learning, specifically the Proximal Policy Optimization (PPO) algorithm~\cite{ppo}, to learn an optimal trajectory tracking strategy. This policy serves as an expert supervisor, providing high-quality demonstration data for the subsequent student training phase. 

The student policy then aims to replicate the teacher's expert performance whilst operating under realistic conditions, learning to fuse unreliable, non-privileged internal state information from the physical dVRK with reliable external visual feedback to achieve precise control. 

However, naive Imitation Learning approaches for this policy distillation suffer from distribution shift. To address this challenge, we employ the Data Aggregation (DAgger) algorithm~\cite{dagger}, an interactive imitation learning method that enhances the student's robustness by iteratively collecting expert corrections. It enables the policy to learn effective recovery strategies from its own errors.

\subsection{``Ideal" Teacher Policy}

\input{figures/pipeline/pipeline}

\textbf{Training Pipeline.} The teacher policy is trained using privileged information, which is available exclusively in the simulator. Specifically, its input state is composed of the robot's true proprioceptive data $S^{Jnt}_{t}$, consisting of joint positions, joint velocities, and the last executed action. Access to this complete and accurate state information is what significantly enhances the efficiency and final performance of the RL training process. The input also contains a preview of 30 future waypoints from the target trajectory $S^{Tgt}_{t:t+n}$, which allows the policy to anticipate upcoming movements for smoother control. The policy itself is formed by a Multi-Layer Perceptron (MLP), and its output is a 6-dimensional vector representing the relative position increment for each joint at the next step.

\input{tables/reward_defination}

\textbf{Reward Design.} The policy is guided by a reward function designed to encourage both high tracking precision and smooth motion. The specific components and their weights are detailed in \autoref{tab:reward_terms_weights}. As position and orientation errors are often difficult to optimize simultaneously, we employ a multiplicative coupling of their respective reward terms. This design compels the policy to be balanced, as it must reduce both errors simultaneously to achieve a high reward. Furthermore, for a high-precision demand such as a surgical task, balancing initial exploration with final fine-grained control is critical. We therefore introduce a curriculum learning strategy that progressively tightens the reward function's error tolerances, effectively guiding the policy from initial, large-scale movements to the delicate, high-precision adjustments required in the final stages. The detailed curriculum schedule is provided in \autoref{tab:curriculum_params}.

\textbf{Fine-Tuning for Robustness.} A teacher that is only an expert on the ideal trajectory (stage 1) is insufficient for guiding an imperfect imitator. During the subsequent imitation learning, the student policy will inevitably explore states that deviate from the expert's distribution. To proactively address this, the teacher undergoes a second stage (stage 2) of fine-tuning with the goal of transforming it from a perfect executor into a ``recovery expert". This is achieved by continuing the training while introducing procedural joint perturbations of gradually increasing magnitude and noise rate. This process forces the policy to learn how to recover from significant state deviations, resulting in a robust teacher that can provide effective, corrective supervision throughout the student's learning process. 

\input{tables/curriculum_setting}

\subsection{``Realistic" Student Policy}

\textbf{State Representation.} The student's observations consist of potentially inaccurate proprioceptive data from dVRK and visual information needed for the policy to learn how to compensate. 
Similar to the teacher policy, this uses joint positions, joint velocities, and the last executed action.
In addition, our method relies on a more lightweight and robust representation of visual information: a set of five 2D keypoints projected from 3D keypoints on the instrument to represent the end-effector's pose as $S^{Vis}_{t-m:t}$. 
Finally, a history of observations is included in the state to allow the policy to implicitly infer system dynamics for smoother control actions.

\textbf{Domain randomisation.} Domain randomisation is employed during training to improve the policy's robustness. 
We note that real dVRK kinematic errors are known to be configuration-dependent, correlated across joints, and history-dependent~\cite{Kinematic_error_learning_based_estimator,Kinematic_error_learning_based_compensator,kinematic_inaccuracy}. Rather than faithfully replicating this error structure, our approach follows the established domain randomisation principle: training under an unstructured distribution that envelops the real-world variation produces robust transferable policies. 
The magnitude of each randomisation parameter was selected based on estimated real-system error bounds and the convergence of the learned policy.
Therefore, independent random biases are sampled uniformly from $\pm0.05$ rad for each joint and held fixed throughout the episode, simulating a consistent but random kinematic offset.
Similarly, to make the policy less sensitive to camera placement, the camera's pose is randomly initialised with a positional offset of up to $\pm2\,cm$ and a rotational offset of up to $\pm10\,\degree$. This encourages the policy to learn view-invariant features and allows for a more flexible real-world setup without requiring precise calibration. Finally, in addition to these per-episode randomisations, a small amount of Gaussian noise is applied to all observation inputs at each time step to further enhance the policy's robustness.

\textbf{Policy Architecture.} Inspired by ACT~\cite{ACT}, a Transformer architecture comprising 4 encoder layers and 1 decoder layer is employed to capture complex temporal and multi-modal dependencies. The self-attention mechanism can explicitly model these dependencies, which is crucial for achieving precise and smooth control. Each proprioceptive state and the keypoint state are tokenised via linear projections, with modality-specific embeddings added to each token. Learnable positional embeddings are added to the token representations of historical observations within a fixed-length temporal window.

\textbf{Training Pipeline.} The student policy is trained to minimize an imitation MSE loss between its predicted action and the teacher's expert action:

\begin{equation}
\label{eq:imitation_loss}
    \mathcal{L} = \mathbb{E}_{s_{t}} \left[ \left\| \hat{a}_{t} - a_{t}^{*} \right\|_{2}^{2} \right],
\end{equation}
where $\hat{a}_{t}$ is the predicted action and $a_{t}^{*}$ is the optimal action produced by the teacher policy.

Training exclusively on new interactive data, however, can cause the policy to catastrophically forget the initial expert behaviour. This is addressed by a dual-buffer strategy that samples from both a static Expert Buffer of ideal trajectories and a rolling Mixed Buffer of data from the DAgger phase~\cite{dagger}. This combination of a structured curriculum and balanced data sampling allows the policy to learn stable error correction while retaining the high precision of the original expert behaviour.

\subsection{Sim-to-Real Transfer}

A successful sim-to-real transfer requires a real-world perception module that can provide 2D keypoint coordinates with high consistency and low latency. We adopt the pyramidal Lucas-Kanade (LK) optical flow algorithm to track the keypoints. In our workflow, an operator first manually initializes keypoints on the initial frame, which are then automatically tracked in real-time by the LK algorithm. This method allows fast and accurate tracking, and provides the geometric consistency over other more granular tracking methods~\cite{cotracker}.

\section{EXPERIMENTS}

In this section, we present a series of experiments to evaluate the effectiveness of our proposed policy. We first introduce the experimental setup and simulation performance, followed by a detailed analysis addressing the following key research questions:

\textbf{Q1.} How effectively does our policy compensate for the dVRK's kinematic inaccuracies when the target trajectory is relocated to different regions of the workspace?

\textbf{Q2.} How well does our policy generalise across different camera viewpoints?

\textbf{Q3.} To what extent does the policy's success depend on the availability and quality of keypoint observations?

\subsection{Environment Setup}

\textbf{Robotic platform.} Our experimental setup is a distributed system designed for real-time, closed-loop control, with the overall architecture illustrated in \autoref{fig:ros}. Teleoperation is performed using a Sigma~7 controller to command the Cartesian motion of the dVRK instrument. The Realsense D405 Camera video stream and the dVRK trajectory information are recorded. All communication between these components is managed by the Robot Operating System (ROS), enabling the entire loop to operate at 30Hz. The resulting action command is executed by the dVRK with a PD controller.

\textbf{Task Descriptions.} A modified peg transfer task was adopted as a benchmark to validate the proposed framework. The standard peg transfer task is widely used as an evaluation benchmark on the dVRK~\cite{Auto_peg_transfer1, Auto_peg_transfer2, Auto_peg_transfer_3}, owing to its clear correspondence to clinical manipulation skills. To increase task complexity and introduce orientation variability, the task objective is defined as placing a peg onto a vertical stick on the pegboard, whilst the peg is initialised on its side adjacent to the board. This setup encourages greater wrist rotation of the instrument, making the task more challenging and representative of dexterous surgical manipulation.

\textbf{Data Collection.} A total of 80 successful demonstrations were collected via teleoperation. Domain randomisation was applied throughout data collection by varying the position and orientation of both the peg and the pegboard on the workspace surface, initialising the robot arm from diverse configurations, and perturbing the camera pose.

\input{figures/ros/ros}

\textbf{Baseline models.} To comprehensively evaluate our proposed policy, its performance is compared against two distinct baselines:
\begin{itemize}
    \item \textbf{IK Replay:} This replays the recorded Cartesian trajectory from teleoperation in joint space relying on the inverse kinematics (IK) of the dVRK. It is a non-learning, open-loop method that is affected by the dVRK's kinematic inaccuracies.
    \item \textbf{ACT:} The Action Chunking Transformer \cite{ACT} is a state-of-the-art imitation-learning-based policy for robotic manipulation. It provides a strong benchmark for comparison with other task-level policy learning methods.
\end{itemize}
The primary evaluation metric for all real-world experiments is the Success Rate.  A trial on the peg-transfer task is considered successful if the peg is transferred from its start position to the target pin in a smooth motion, without getting stuck.

Training was carried with the ORBIT-Surgical framework~\cite{yu2024orbit}, which is built upon NVIDIA Isaac Sim, using a single NVIDIA RTX 5090 GPU. The two-stage teacher policy training took approximately 40 hours in total, utilizing 8192 parallel environments for data collection, with a total batch size of $8192\,\times\,64$. The student policy was subsequently trained for approximately 20 hours using 200 parallel environments and a batch size of $200\,\times\,256$. The ACT was trained on the demonstration dataset for 200k gradient steps with its original hyper-parameters. The proprioceptive modalities are tokenised identical to the proposed policy while visual inputs are tokenised respective to the original implementation. 

\subsection{Simulation Performance}

Before deployment on the physical robot, the trained policies were first validated in the simulation environment. This evaluation serves two main purposes: to verify the effectiveness of our teacher-student learning framework and to establish an upper-bound performance benchmark for the subsequent real-world experiments.

While the teacher policy achieves near-perfect trajectory tracking with privileged information, the student policy exhibits slight but expected deviations. To precisely quantify these observations, we present a summary of the average performance in \autoref{tab:simulation_errors}. The visualisation of the error through a rollout episode is shown in \autoref{fig:step_tracking_error}. This shows that the teacher policies maintain a position error and orientation of approximately $0.6 \, mm$ and $0.015 \, rad$ throughout the trajectory. The student policy error, while slightly higher at around $1.2 \, mm$ and $0.03 \, rad$, remains low and stable, following the teacher's performance profile. Notably, the Stage 2 teacher's performance is nearly identical to that of Stage 1, demonstrating that the robustness fine-tuning did not sacrifice precision.

\input{figures/dVRK/step_err}

The graph in \autoref{fig:step_tracking_error} indicates a slight upward trend in student error, which is an acceptable outcome for DAgger. This is because while DAgger mitigates the cumulative error arising from distribution shifts during imitation learning, it cannot entirely eliminate it. 
These results quantitatively confirm the success of our proposed learning framework. The student policy successfully distils the precision tracking ability from the expert teacher, which learns to achieve high-precision tracking by leveraging external visual feedback to compensate for its inaccurate proprioceptive information.

\input{tables/avg_err}

\subsection{Real-World Deployment}

\subsubsection{Relocation generalizability}

Relocation generalizability experiments were conducted to evaluate the policy's effectiveness across different workspace configurations. In this experimental protocol, the workspace was horizontally shifted relative to the robot whilst maintaining the internal relative positioning of objects. The policy was required to complete the task by replicating the demonstration trajectory within the shifted workspace configuration.

This experiment could test policy's ability to compensate for the dVRK's kinematic inaccuracies across different regions of the workspace. 
The underlying hypothesis is that these inaccuracies are workspace-dependent, as different joint configurations can lead to varying error characteristics. A truly robust policy must therefore generalize its corrective behaviour rather than overfit to a single region.

To test this, we evaluate our policy against a traditional open-loop inverse kinematic baseline, named as inverse kinematic (IK) Replay, which is supposed to achieve the task with high success rate when there are no kinematic issues. The expert trajectories for the peg-transfer task were systematically relocated horizontally (along the x-axis) from $ -4\,cm $ to $ +6\,cm$, forcing the robot to operate in different joint configurations within its workspace. 

\input{tables/experiment1}

\input{figures/relocation/generalization}

The results of this experiment are summarized in \autoref{tab:relocation}. The IK Replay baseline performs well in a very narrow range, achieving 100\% success at the original location (0 cm) and with a small ±2 cm offset. However, its performance collapses to 0\% success with any larger offset. In contrast, our policy demonstrates significantly more robust performance. It maintains a high success rate of over 80\% across a wide range of offsets (±2 cm). Even at the largest tested offset of +6 cm, it still achieves a remarkable success rate of 55\% compared to IK Replay.

The failure of IK Replay under large workspace offsets can be attributed to the dVRK's kinematic inaccuracies. The initial success at the original position is deceptive; it occurs not because the system is inherently accurate, but because the teleoperated trajectory was collected in this exact configuration and has implicitly encoded the specific, configuration-dependent kinematic errors of that workspace region. This open-loop method is fragile because the robot's joint configuration and its corresponding error profile change when the trajectory is relocated to different workspace positions. Furthermore, the consistent failure pattern highlights that the underlying issue represents a repeatable, deterministic kinematic error rather than random control noise.

In contrast to the brittle IK Replay, our policy's success demonstrates the key advantage of its learning-based design. Instead of relying on a flawed internal kinematic model, it has learned a generalizable mapping from visual error to corrective action. This allows it to handle the robot's configuration-dependent inaccuracies, allowing it to succeed where the open-loop method fails.

Despite its overall performance, we also observe that our policy's success rate presents a slight downward trend as the trajectory offset increases. We attribute this to a combination of three potential factors. Firstly, the expert dataset may not fully cover the entire workspace, causing the policy to encounter out-of-distribution states at large offsets. Secondly, a sim-to-real gap in perception may arise. While keypoints are derived from ideal projections in simulation, the physical camera's lens distortion can reduce tracking accuracy near the periphery of the image. Thirdly, the largest offsets may push the robot towards its physical workspace boundaries, where its mechanical control and precision can be inherently less reliable.

\subsubsection{Viewpoint Generalizability}

A key objective of our work is to develop a policy that is robust to variations in camera placement, relaxing the strict requirement for a fixed, precisely calibrated camera common in many SOTA methods. To quantitatively evaluate this generalizability, we compare our policy against another baseline, the Action Chunking Transformer~\cite{ACT}. The camera pose was systematically varied from its initial calibrated position, with perturbations in both translation (±1 cm and ±3 cm) and rotation (±5° and ±10°).

\input{tables/experiment2}

The results of this comparison are presented in \autoref{tab:viewpoint}. When operating at original calibrated camera pose, both our policy and the ACT baseline achieve a high success rate of 95\%. As the camera pose is varied, the performance of both methods decreases, but at different rates. Under the largest translational shift of ±3 cm, our policy's success rate is 65\%, while ACT's is 35\%. Similarly, under the largest rotational variation of ±10°, our policy achieves a 40\% success rate, compared to 15\% for the ACT baseline. The data shows that our policy maintains a consistently higher success rate than the ACT baseline under the camera pose perturbations.

The results reveal that the superior generalisability of our policy is a direct outcome of our training methodology. The camera pose randomisation employed during training forced our policy to learn a more view-invariant control strategy, focusing on geometric relationships that remain consistent across different viewpoints. In contrast, the ACT collapses when the visual input shifts into an out-of-distribution domain. However, the results also show a degradation in our policy's performance as the perturbations become larger.

\subsubsection{Visual Features Robustness}
Reliable keypoint tracking in real surgical environments remains challenging; hence, robustness to corrupted or incomplete observations is a critical property for any clinically viable policy. To evaluate this, we conduct an ablation study examining two complementary aspects of the policy's dependence on external keypoint inputs: (1) its tolerance to observation corruption, and (2) the necessity of keypoint feedback beyond proprioceptive information alone. Results are presented in \autoref{tab:tracking}.

\textbf{Robustness to corrupted keypoint observations.} In \textbf{Noisy}, Gaussian noise with a standard deviation of 5 pixels was added to all keypoint estimates at inference time. The policy maintained robust performance, with the success rate declining only marginally from $90\%$ to $80\%$, demonstrating tolerance to moderate localisation errors. In the \textbf{Drop 2} condition, two of the five tracked keypoints were randomly masked during each episode of inference, causing the success rate to drop substantially to $30\%$. These results indicate its inherent tolerance to minor tracking jitter commonly encountered in real-world keypoint tracking systems.

\textbf{Necessity of keypoint feedback over proprioception alone.} When four keypoints were masked (\textbf{Drop 4}), the policy failed entirely, yielding a $0\%$ success rate. In the \textbf{No Access} condition, the keypoint modality was entirely removed during both training and inference, yielding a $0\%$ success rate. This confirms that the policy does not succeed by relying solely on inaccurate proprioceptive signals. Instead, it utilised keypoint-based visual feedback to perceive and adapt to the current pose.

\input{tables/experiment3}

\section{DISCUSSION AND CONCLUSION}\label{discussion}

This paper presents a novel visuomotor controller developed within a teacher-student learning framework, that actively compensates for the inherent kinematic inaccuracies of surgical robots by leveraging closed-loop visual feedback. An expert teacher, trained in an ideal simulation via reinforcement learning with privileged information, provides supervision for a student policy that learns through an interactive process to fuse unreliable proprioception with reliable external 2D keypoints. Real-world deployment of the learned policy demonstrated its ability to compensate for workspace-dependent kinematic errors and its significant robustness to camera pose variations. Therefore, this work establishes a complete methodology for developing low-level controllers that can overcome physical hardware limitations for precise surgical trajectory tracking.

However, this work still faces key challenges. Firstly, the visual perception module has critical limitations: it requires manual initialization and is fragile to visual perturbations like keypoint occlusion and significant lighting changes; moreover, the projective nature of 2D keypoints can make it difficult to unambiguously represent the tool's complete 3D pose. Secondly, whilst simulated linear errors are incorporated during model training, non-linear systematic errors present in the dVRK are not modelled, potentially limiting the framework's capacity to compensate for such disturbances. Thirdly, the policy has only been exhaustively evaluated on a single exemplar task, which may not fully reflect its performance across the broader spectrum of surgical procedures.

Future work should focus on several key directions. We intend to integrate an autonomous perception system capable of learning to select task-relevant keypoints whilst generalising to novel surgical tools and environments. Additionally, non-linear dVRK system errors will be characterised and injected into the simulation to enable the policy to compensate for such disturbances more effectively. The proposed framework will further be evaluated across a broader range of surgical tasks to assess its generalisation capability and robustness. We hope this work serves as a foundational step towards overcoming the hardware limitations of current surgical robots, paving the way for more robust and intelligent surgical automation systems.


\section*{ACKNOWLEDGMENTS}
This paper is independent research funded by the National Institute for Health Research (NIHR) Imperial Biomedical Research Centre (BRC), the Cancer Research UK (CRUK) Imperial Centre, the Wellcome Trust ITPA MedTechOne awards. 

\bibliographystyle{IEEEtran}
\bibliography{icra_26}

\end{document}

%% file: figures/graphical_abstract/graphical_abstract.tex
\twocolumn[{%
	\renewcommand\twocolumn[1][]{#1}%
	\maketitle
        \vspace{-4mm}
	\begin{center}
        \includegraphics[width=\textwidth]{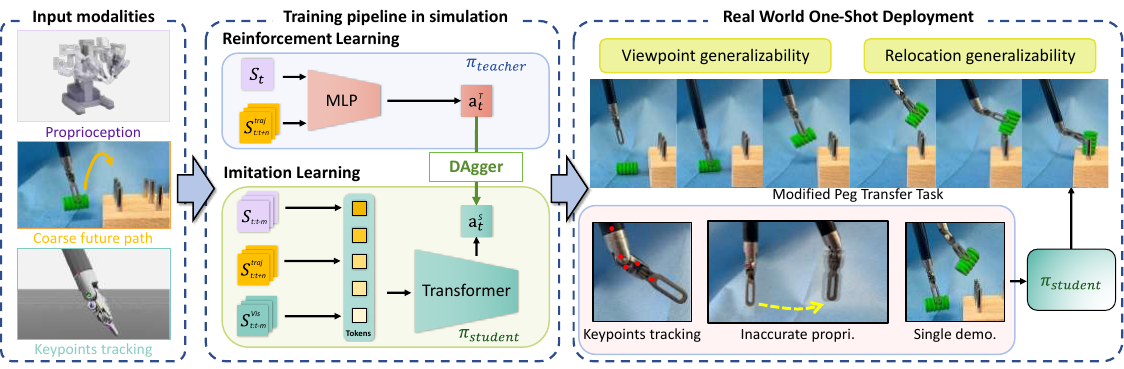}
        \captionof{figure}{ {\label{fig:graphical_abstract}
        We proposed a learning framework that actively compensates for a surgical robot's kinematic inaccuracies by training a visuomotor controller within a teacher-student paradigm. The policy learns to fuse unreliable proprioceptive data with reliable visual feedback, enabling robust generalization to variations in camera viewpoint and workspace relocation upon sim-to-real deployment to a physical dVRK.
        }}

	\end{center}
}]

%% file: figures/pipeline/pipeline.tex
\begin{figure*}[!t]
\centering
\includegraphics[width=\linewidth]{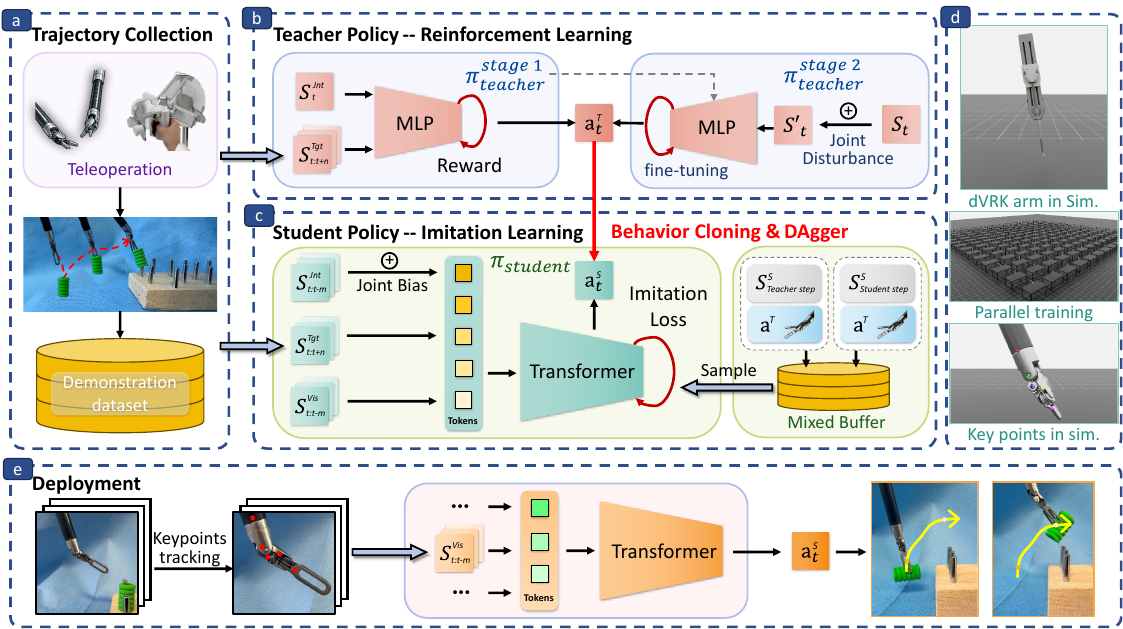}
\caption{\textbf{The overview of the Training framework}: \textbf{(a)}: The framework begins with collecting expert trajectories via teleoperation. \textbf{(b)}: A teacher policy is then trained in simulation using reinforcement learning and fine-tuned with joint disturbances to develop robust recovery behaviour. \textbf{(c)}: Subsequently, a student policy uses Imitation Learning (DAgger) to distil the teacher's expertise, learning to fuse biased proprioceptive data with external visual keypoints to compensate for kinematic inaccuracies introduced in simulation. \textbf{(d)}: The simulation comprises multiple parallel training environments, each containing one dVRK patient-side arm. Keypoints tracked in the simulation are indicated with colour labels. \textbf{(e)}: Finally, the trained student policy is deployed directly on the physical robot for real-world task execution.}
\label{fig:pipeline}
\end{figure*}

%% file: tables/reward_defination.tex
\begin{table}[b]
    \centering
    \caption{Reward Function Terms and Weights}
    \label{tab:reward_terms_weights}

    \renewcommand{\arraystretch}{1.5}

    \begin{tabular}{ccc}
        \toprule
        \textbf{Reward Terms} & \textbf{Expressions} & \textbf{Weights} \\
        \midrule
        Position error & $ \exp\left(-\frac{\|\mathbf{p} - \mathbf{p}^{\text{tg}}\|^2}{\sigma^2_{\text{pos}}}\right) $ & 0.6 \\
        Orientation error & $ \exp\left(-\frac{\|\mathbf{q} - \mathbf{q}^{\text{tg}}\|^2}{\sigma^2_{\text{orn}}}\right) $ & 0.1 \\
        Action rate & $ -\|\dot{\mathbf{a}}\|^2 $ & 0.35 \\
        Joint velocities & $ -\|\dot{\mathbf{\theta}}\|^2 $ & 0.01 \\
        Balanced error & $ \exp\left(-\left(\frac{\|\mathbf{p} - \mathbf{p}^{\text{tg}}\|^2}{\sigma^2_{\text{pos}}} + \frac{\|\mathbf{q} - \mathbf{q}^{\text{tg}}\|^2}{\sigma^2_{\text{orn}}}\right)\right) $ & 8 \\
        \bottomrule
    \end{tabular}
\end{table}

%% file: tables/curriculum_setting.tex
\newcolumntype{C}{>{\centering\arraybackslash}X}

\begin{table}[tb]
    \centering
    \caption{Curriculum Learning Parameters}
    \label{tab:curriculum_params}

    \begin{threeparttable}
        \begin{tabularx}{\linewidth}{CCCC}
            \toprule
            \multicolumn{2}{c}{\textbf{Position Error Curriculum}} & \multicolumn{2}{c}{\textbf{Orientation Error Curriculum}} \\
            \cmidrule(r){1-2} \cmidrule(l){3-4}

            \textbf{$\epsilon_{pos}$} & \textbf{$\sigma^2_{pos}$} & \textbf{$\epsilon_{orn}$} & \textbf{$\sigma^2_{orn}$} \\
            \midrule
            0.3 & 0.13 & 1.0 & 1.44 \\
            0.15 & 0.032 & 0.6 & 0.52 \\
            0.08 & 0.0092 & 0.15 & 0.032 \\
            0.01 & 0.00014 & 0.08 & 0.0092 \\
            0.005 & $3.6 \times 10^{-5}$ & 0.05 & $3.6 \times 10^{-3}$ \\
            0.003 & $1.3 \times 10^{-5}$ & 0.03 & $1.3 \times 10^{-3}$ \\
            0.002 & $5.76 \times 10^{-6}$ & 0.02 & $5.76 \times 10^{-4}$ \\
            0.001 & $1.44 \times 10^{-6}$ & 0.01 & $1.44 \times 10^{-4}$ \\
            \bottomrule
        \end{tabularx}

        \begin{tablenotes}[flushleft]
            \footnotesize
            \item For position error curriculum, $\epsilon_{pos}$ (in m) and $\sigma^2_{pos}$ represent the threshold and sigma in the reward term. For orientation error curriculum, $\epsilon_{orn}$ (in rad) and $\sigma^2_{orn}$ represent the threshold and sigma in the reward term.
        \end{tablenotes}

    \end{threeparttable}
\end{table}

%% file: figures/ros/ros.tex
\begin{figure}[t]
\includegraphics[width=\linewidth]{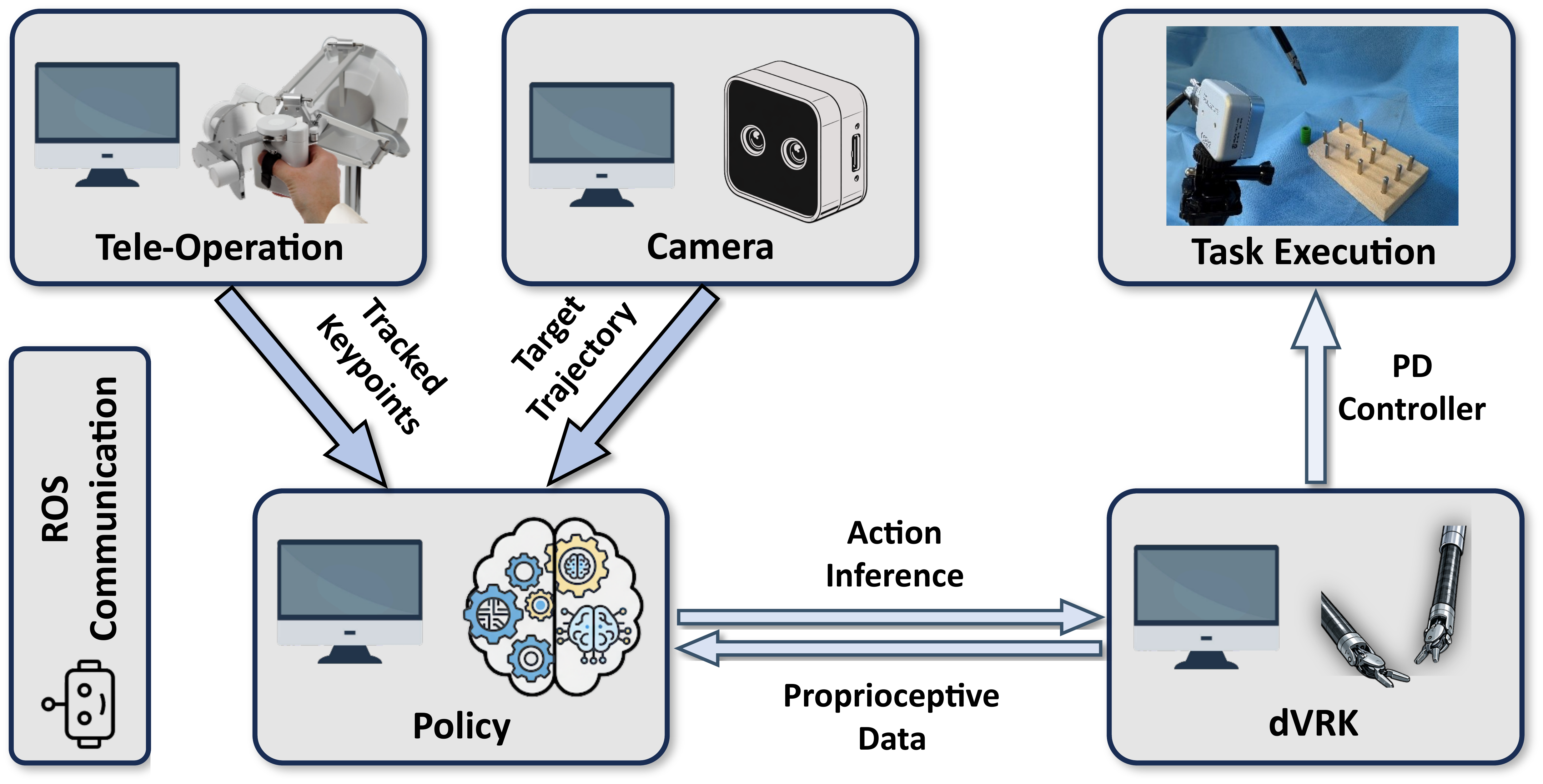}
\caption{\textbf{Overview of the robotic control system}: The central Policy module integrates a target trajectory from the teleoperation system, tracked keypoints from a camera, and proprioceptive data from the dVRK to infer the action. This inferred action is then sent back to the dVRK, whose internal PD Controller performs the final task execution. The entire distributed system is connected and communicates via ROS. }
\label{fig:ros}
\end{figure}

%% file: figures/dVRK/step_err.tex
\begin{figure}[tbp]
    \centering
    \begin{minipage}{0.49\linewidth}
        \centering
        \includegraphics[width=\linewidth]{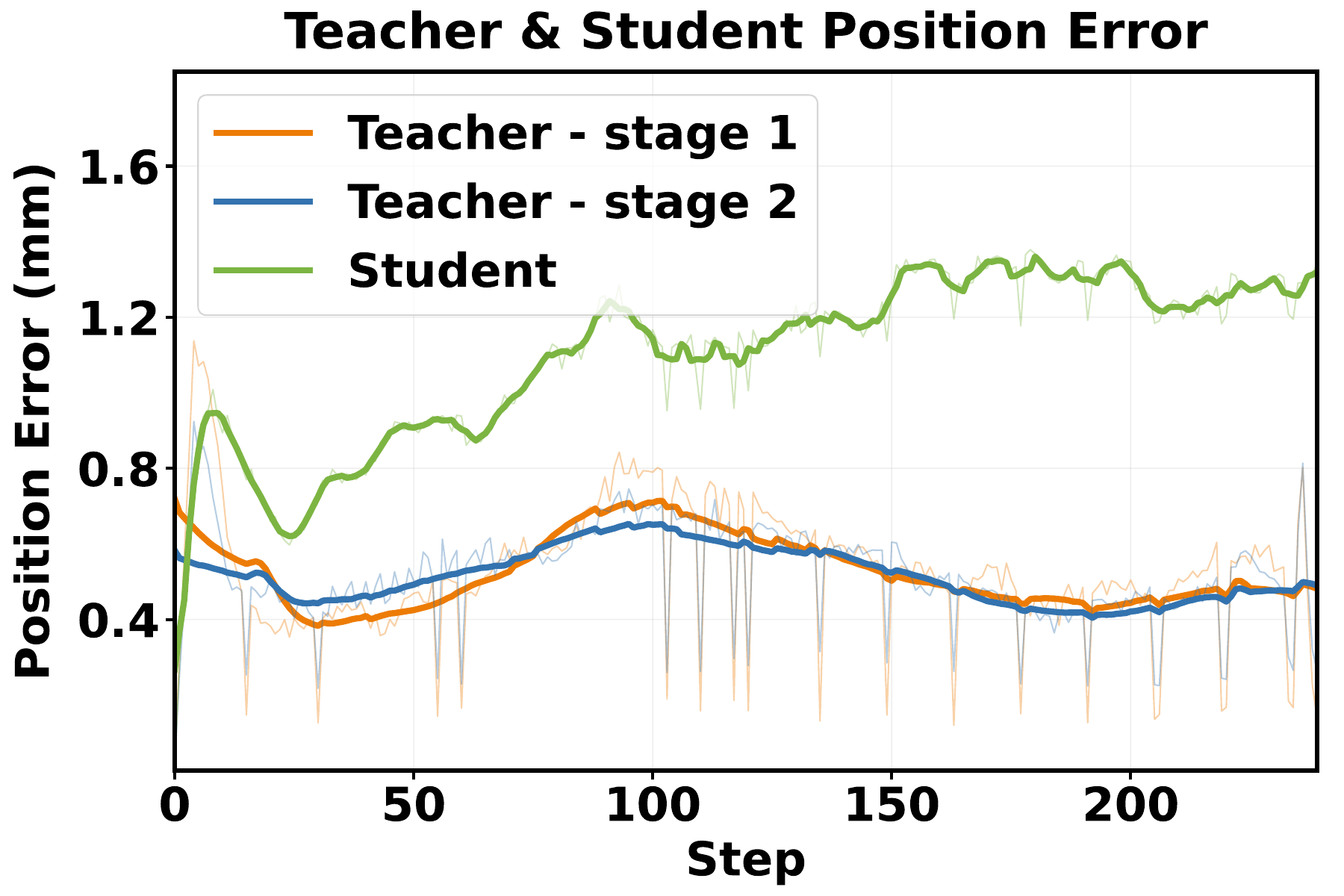}
    \end{minipage}
    \hfill
    \begin{minipage}{0.49\linewidth}
        \centering
        \includegraphics[width=\linewidth]{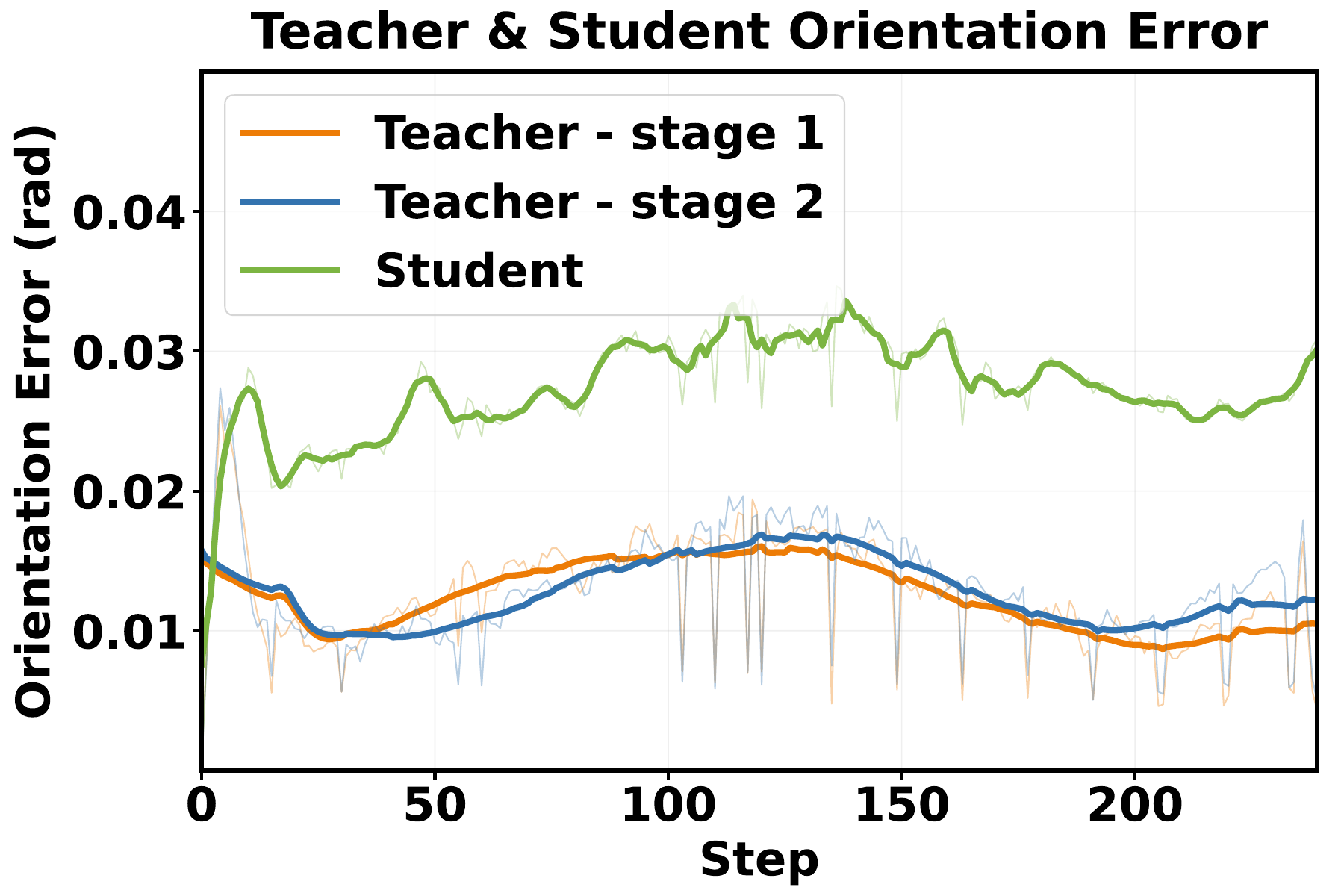}
    \end{minipage}

    \caption{\textbf{Tracking Performance of Teacher and Student Policies in Simulation.} The plots compare the position error (\textbf{left}) and orientation error (\textbf{right}) for the teacher and student policies over a rollout trajectory. Both Stage 1 and Stage 2 teacher policies demonstrate nearly identical, high-precision performance with low, stable errors. The student policy successfully tracks the trajectory but exhibits a consistently higher tracking error.}
    \label{fig:step_tracking_error}
\end{figure}

%% file: tables/avg_err.tex
\begin{table}[b]
    \centering
    \caption{Tracking Error in Simulation.}
    \label{tab:simulation_errors}

    \begin{tabular}{l ccc}
        \toprule
        \multirow{2}{*}{\textbf{Metric}} & \multicolumn{2}{c}{\textbf{Teacher Policy}} & \multirow{2}{*}{\textbf{Student Policy}} \\
        \cmidrule(lr){2-3}
         & \textbf{Stage 1} & \textbf{Stage 2} & \\
        \midrule
        Position error [$mm$] & 0.6 & 0.6 & 1.2 \\
        Orientation error [$rad$] & 0.015 & 0.015 & 0.03 \\
        \bottomrule
    \end{tabular}
\end{table}

%% file: tables/experiment1.tex
\begin{table}[tbp]
    \centering
    \caption{Relocation Robustness}
    \label{tab:relocation}
\begin{threeparttable} 
    \begin{tabular}{l cccccc}
        \toprule
        & \multicolumn{6}{c}{$\uparrow$\textbf{Success Rate after  $x_{shift}\,[cm]$}} \\
        \cmidrule(lr){2-7}
        \textbf{Method} & \textbf{-4} & \textbf{-2} & \textbf{0} & \textbf{+2} & \textbf{+4} & \textbf{+6} \\
        \midrule
        \textbf{Ours} & \textbf{15/20} & 17/20 & 18/20 & 19/20 & \textbf{15/20} & \textbf{11/20} \\
        \textbf{IK Replay} & 0/20 & \textbf{20/20} & \textbf{20/20} & \textbf{20/20} & 0/20 & 0/20 \\
        \bottomrule
    \end{tabular}
\begin{tablenotes}[flushleft]   
\footnotesize
\item The trajectory was relocated by shifting $x_{shift}$ it horizontally on the platform in a range of $-4\,cm$ to $6\,cm$, where $0\,cm$ represents the original location. \textbf{Ours} represents our proposed method.
\end{tablenotes}

\end{threeparttable}
\end{table}

%% file: figures/relocation/generalization.tex
\begin{figure*}[!t]
\centering
\includegraphics[width=\linewidth]{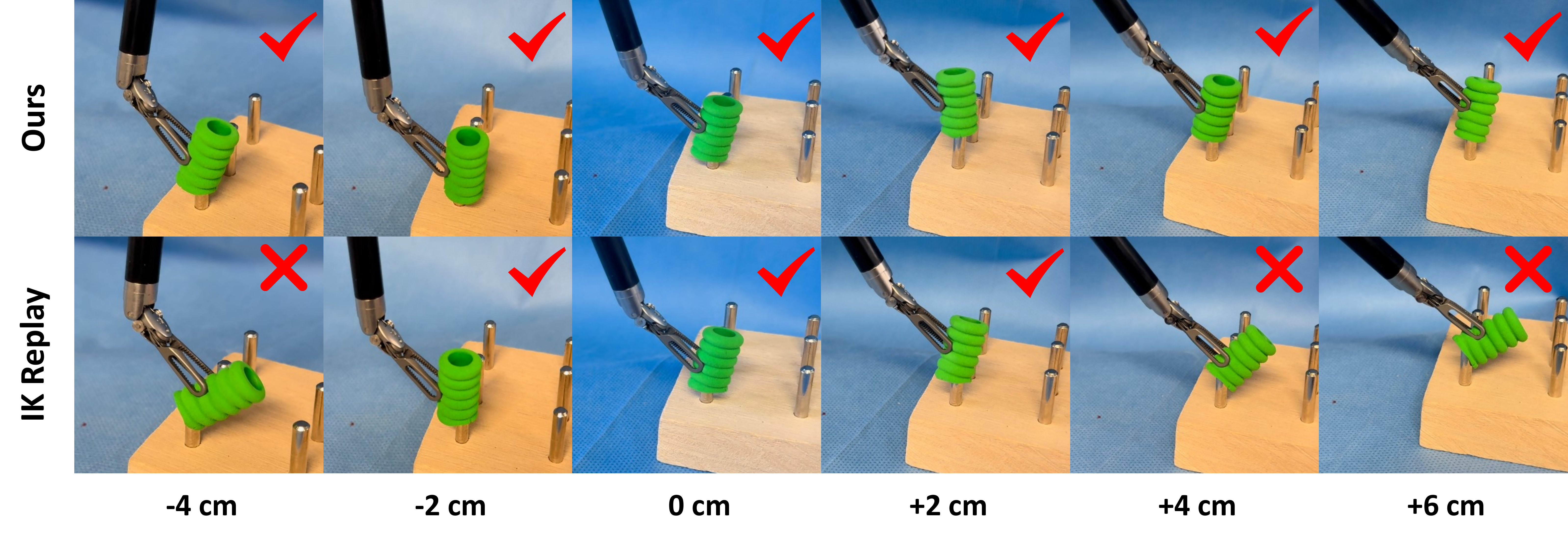}
\caption{\textbf{Example frames from the Relocation Generalizability experiment.} This figure presents a qualitative comparison between our policy (\textbf{Ours}, top row) and the \textbf{IK Replay} baseline (bottom row). The peg transfer task was systematically shifted horizontally from -4 cm to +6 cm. The results show that our policy successfully completes the task across a wide range of locations (indicated by \cmark). In contrast, the IK Replay baseline succeeds only at or near the original location (0 cm) but fails at larger offsets (indicated by \xmark). \textbf{Ours} represents our proposed method.}
\label{fig:relocation}
\end{figure*}

%% file: tables/experiment2.tex
\begin{table}[tbp]
    \centering
    \caption{Viewpoint Robustness}
    \label{tab:viewpoint}
\begin{threeparttable} 
\centering
\setlength{\tabcolsep}{10pt}
\resizebox{\columnwidth}{!}{
    \begin{tabular}{l ccccc}
        \toprule
        & \multicolumn{5}{c}{\textbf{$\uparrow$Success Rate under
        Camera Pose Variation}} \\
        \cmidrule(lr){2-6}
        \textbf{Method} & \textbf{Original} & \textbf{±1 cm} & \textbf{±3 cm} & \textbf{±5°} & \textbf{±10°} \\
        \midrule
        \textbf{Ours} & 19/20 & \textbf{17/20} & \textbf{13/20} & \textbf{14/20} & \textbf{8/20} \\
        \textbf{ACT} & 19/20 & 15/20 & 7/20 & 10/20 & 3/20 \\
        \bottomrule
    \end{tabular}
}%

\begin{tablenotes}[flushleft]   
\footnotesize
\vspace{3pt}
\item \parbox{\columnwidth}{During the evaluation, the camera pose was varied to assess its sensitivity to pose changes. Two types of variation were considered: horizontal translation (in $cm$) and yaw rotation (in $\degree$). \textbf{Original} indicated the original camera pose identical to the training setup.}
\end{tablenotes}

\end{threeparttable}
\end{table}

%% file: tables/experiment3.tex
\begin{table}[htbp]
    \centering
    \caption{Robustness to Keypoints Tracking}
    \label{tab:tracking}
\begin{threeparttable} 
\centering
    \begin{tabular}{l cccccc}
        \toprule
        & \multicolumn{5}{c}{$\uparrow$\textbf{Success Rate}} \\
        \cmidrule(lr){2-6}
        \textbf{Method} & \textbf{Original} & \textbf{Noisy} & \textbf{Drop 2} & \textbf{Drop 4} & \textbf{No access} \\
        \midrule
        \textbf{Ours} & 9/10 & 8/10 & 3/10 & 0/10 & 0/10\\
        \bottomrule
    \end{tabular}

\begin{tablenotes}[flushleft]   
\footnotesize
\item In this ablation study of visual keypoints tracking input, the success rate of five cases are reported. \textbf{Original} is identical to the proposed policy setup. \textbf{Noisy} represents adding Gaussian noise to the tracking input. \textbf{Drop 2} and \textbf{Drop 4} represents masking out two or four out of totally five tracked keypoints. \textbf{No access} means removing this input modality during training and further evaluation
\end{tablenotes}

\end{threeparttable}
\end{table}